# Learning Riemannian Metrics

Guy Lebanon
School of Computer Science
Carnegie Mellon University
Pittsburgh, PA, USA

## Abstract

We consider the problem of learning a Riemannian metric associated with a given differentiable manifold and a set of points. Our approach to the problem involves choosing a metric from a parametric family that is based on maximizing the inverse volume of a given dataset of points. From a statistical perspective, it is related to maximum likelihood under a model that assigns probabilities inversely proportional to the Riemannian volume element. We discuss in detail learning a metric on the multinomial simplex where the metric candidates are pull-back metrics of the Fisher information under a continuous group of transformations. When applied to documents, the resulting geodesics resemble, but outperform, the TFIDF cosine similarity measure in classification.

## 1 Introduction

Machine learning algorithms often require an embedding of data points into some space. Algorithms such as $k$-nearest neighbors and neural networks assume the embedding space to be $\mathbb{R}^n$ while SVM and other kernel methods embed the data in a Hilbert space through a kernel operation.

Whatever the embedding space is, the notion of metric structure has to be carefully considered. The popular assumption of a Euclidean metric structure is often used without justification by data or modeling arguments.

We argue that in the absence of direct evidence of Euclidean geometry, the metric structure should be inferred from data. After obtaining the metric structure, it may be passed to a learning algorithm for use in tasks such as classification and clustering.

Several attempts have recently been made to learn the metric structure of the embedding space from a given data set. Saul and Jordan use geometrical arguments to learn optimal paths connecting two points in a space [12]. Xing et al. [13] learn a global metric structure. Such a metric structure is able to capture non-Euclidean geometry, but only in a restricted manner since the metric is constant throughout the space. Lanckriet et al. [7] learn a kernel matrix that represents similarities between all pairs of the supplied data points. While such an approach does learn the kernel structure from data, the resulting Gram matrix does not generalize to unseen points.

Learning a Riemannian metric is also related to finding a lower dimensional representation of a dataset. Work in this area includes linear methods such as principal component analysis and nonlinear methods such as spherical subfamily models [4] or locally linear embedding [10] and curved multinomial subfamilies [5]. Once such a submanifold is found, distances $d(x,y)$ may be computed as the lengths of shortest paths on the submanifold connecting $x$ and $y$. As shown in Section 2, this approach is a limiting case of learning a Riemannian metric for the embedding high-dimensional space.

Lower dimensional representations are useful for visualizing high dimensional data. However, these methods assume strict conditions that are often violated in real-world, high dimensional data. The obtained submanifold is tuned to the training data and new data points will likely lie outside the submanifold due to noise. It is necessary to specify some way of projecting the off-manifold points into the manifold. There is no notion of non-Euclidean geometry outside the submanifold and if the estimated submanifold does not fit current and future data perfectly, Euclidean projections are usually used.

Another source of difficulty is estimating the dimension of the submanifold. The dimension of the submanifold is notoriously hard to estimate in high dimensional sparse datasets. Moreover, the data may have different lower dimensions in different locations or may lie on several disconnected submanifolds thus violating the assumptions underlying the submanifold approach.

We propose an alternative approach to the metric learning problem. The obtained metric is local, thus capturing local variations within the space, and is defined on the entire embedding space. A set of metric candidates is represented as a parametric family of transformations, or equivalently as a parametric family of statistical models and the obtained metric is chosen from it based on some performance criterion.

In Section 2 we discuss our formulation of the Riemannian metric problem. Section 3 describes the set of metric candidates as pull-back metrics of a group of transformations. Section 4 demonstrates the framework in the case of the multinomial simplex, followed by a discussion of the resulting generative model in Section 5. In Section 6 we apply the framework to text classification and report experimental results on the WebKB data. We conclude with a summary in Section 7.

## 2 The Metric Learning Problem

We start with a brief discussion of some basic concepts from differential geometry and refer to [1] for a more detailed description. A Riemannian metric $g$, on an $n$th dimensional differentiable manifold $\mathcal{M}$, is a function that assigns for each point of the manifold $x \in \mathcal{M}$ an inner product on the tangent space $T_x\mathcal{M}$. The metric is required to satisfy the usual inner product properties and to be $C^\infty$ in $x$.

The metric allows us to measure lengths of tangent vectors $v \in T_x\mathcal{M}$ as $\|v\|_x = \sqrt{g_x(v,v)}$, leading to the definition of a length of a curve on the manifold $c : [a,b] \to \mathcal{M}$ as $\int_a^b \|\dot{c}(t)\| dt$. The geodesic distance function $d(x,y)$ for $x,y \in \mathcal{M}$ is defined as the length of the shortest curve connecting $x$ and $y$ and turns the manifold into a metric space.

The metric learning problem may be formulated as follows. Given a differentiable manifold $\mathcal{M}$ and a dataset $D = \{x_1, \ldots, x_N\} \subset \mathcal{M}$, choose a Riemannian metric $g$ from a set of metric candidates $\mathcal{G}$. As in statistical inference, $\mathcal{G}$ may be a parametric family

$$\mathcal{G} = \{g_\theta : \theta \in \Theta \subset \mathbb{R}^k\} \quad (1)$$

or as in nonparametric statistics a less constrained set of candidates. We focus on the parametric approach, as we believe it to generally perform better in high dimensional sparse data such as text documents.

We propose to choose the metric based on maximizing the following objective function $\mathcal{O}(g, D)$

$$\mathcal{O}(g, D) = \prod_{i=1}^N \frac{\text{dvol} g^{-1}(x_i)}{\int_\mathcal{M} \text{dvol} g^{-1}(x) dx} \quad (2)$$

where $\text{dvol} g(x) = \sqrt{\det g(x)}$ is the differential volume element at the point $x$ according to the metric $g$. Note that $\det g > 0$ since $g$ is positive definite.

The volume element $\text{dvol}(x)$ summarizes the size of the metric at $x$ in a scalar. Intuitively, paths crossing areas with high volume will tend to be longer than the same paths over an area with low volume. Hence maximizing the inverse volume in (2) will result in shorter curves across densely populated regions of $\mathcal{M}$. As a result, the geodesics will tend to pass through densely populated regions. This agrees with the intuition that distances between data points should be measured on the lower dimensional data submanifold, thus capturing the intrinsic geometrical structure of the data.

The normalization in (2) is necessary since the problem is clearly unidentifiable without it. Metrics $cg$ with $0 < c < 1$ will always a have higher inverse volume element than $g$. The normalized inverse volume element may be seen as a probability distribution over the manifold. As a result, we may cast the problem of maximizing $\mathcal{O}(g, D)$ as a maximum likelihood problem.

If $\mathcal{G}$ is completely unconstrained, the metric maximizing the above criterion will have a volume element tending to 0 at the data points and $+\infty$ everywhere else. Such a solution is analogous to estimating a distribution by an impulse train at the data points and 0 elsewhere (the empirical distribution). As in statistics we avoid this degenerate solution by restricting the set of candidates $\mathcal{G}$ to a small set of relatively smooth functions.

The case of extracting a low dimensional submanifold (or linear subspace) may be recovered from the above framework if $g \in \mathcal{G}$ is equal to the metric inherited from the embedding Euclidean space across a submanifold and tending to $+\infty$ outside. In this case distances between two points on the submanifold will be measured as the shortest curve on the submanifold using the Euclidean length element.

If $\mathcal{G}$ is a parametric family of metrics $\mathcal{G} = \{g_\lambda : \lambda \in \Lambda\}$, the log of the objective function $\mathcal{O}(g)$ is equivalent to the loglikelihood $\ell(\lambda)$ under the model

$$p(x; \lambda) = \frac{1}{Z}(\sqrt{\det g_\lambda(x)})^{-1}.$$

Such a model is the inverse of Jeffreys' prior $p(x) \propto \sqrt{\det g(x)}$. However in the case of Jeffreys' prior, the metric is known in advance and there is no need for parameter estimation. For prior work on connecting volume elements and densities on manifolds refer to [9].

Specifying the family of metrics $\mathcal{G}$ is not an intuitive task. Metrics are specified in terms of a local inner product and it may be difficult to understand the implications of a specific choice on the resulting distances. The next section describes an intuitive way of specifying the family $\mathcal{G}$ as pull-back metrics of a set of transformations.

## 3 Pull-back Metrics and Flattening Transformations

Let $F : \mathcal{M} \to \mathcal{N}$ be a diffeomorphism of the manifold $\mathcal{M}$ onto the manifold $\mathcal{N}$. Let $T_x\mathcal{M}, T_y\mathcal{N}$ be the tangent spaces to $\mathcal{M}$ and $\mathcal{N}$ at $x$ and $y$ respectively. Associated with $F$ is the push-forward map $F_*$ that maps $v \in T_x\mathcal{M}$ to $v' \in T_{F(x)}\mathcal{N}$. It is defined as

$$v(h \circ F) = (F_*v)h, \quad \forall h \in C^\infty(\mathcal{N}).$$

Intuitively, the push forward maps velocity vectors of curves to velocity vectors of the transformed curves.

Assuming a Riemannian metric $h$ on $\mathcal{N}$, we can obtain a metric $F^*h$ on $\mathcal{M}$ called the pullback metric

$$F^*h_x(u,v) = h_{F(x)}(F_*u, F_*v)$$

where $F_*$ is the push-forward map defined above. The importance of this map is that it turns $F$ (as well as $F^{-1}$) into an isometry; that is,

$$d_{F^*h}(x,y) = d_h(F(x), F(y)).$$

Consider the case were the data $D \subset \mathcal{M} = \mathcal{N}$ and $h$ is the Fisher information metric. Instead of specifying a parametric family of metrics as discussed in the previous section, we specify a parametric family of transformations $\{F_\lambda : \lambda \in \Lambda\}$. The resulting set of metric candidates will be the pull-back metrics $\mathcal{G} = \{F_\lambda^* h : \lambda \in \Lambda\}$.

If $\mathcal{N} \subset \mathbb{R}^n$, $h$ is the metric inherited from the Euclidean metric in $\mathbb{R}^n$ and $D \subset \mathcal{M}$, we call $F$ a flattening transformation. Distances on the manifold $(\mathcal{M}, F^*h)$ may be measured as the shortest Euclidean path on the manifold $\mathcal{N}$ between the transformed points. $F$ thus takes a locally distorted space and converts it into a subset of $\mathbb{R}^n$ with the metric inherited from the Euclidean embedding space.

In the next sections we work out in detail an implementation of the above framework in which the manifold $\mathcal{M}$ is the multinomial simplex.

## 4 The Multinomial Simplex

We now apply the metric learning framework to the case of the $n$-simplex $\mathcal{P}_n$, defined by

$$\mathcal{P}_n = \left\{ x \in \mathbb{R}^{n+1} : \forall i, \ x_i \geq 0, \ \sum_{i=1}^{n+1} x_i = 1 \right\}.$$

Every point in $\mathcal{P}_n$ corresponds to a multinomial model over $n+1$ possible outcomes. The coordinates $\{x_i\}$ describe the probability of obtaining different outcomes in a single experiment. We refer to the interior of the simplex as $\text{int}\mathcal{P}_n$.

The Fisher information metric on $\mathcal{P}_n$ is given by

$$\mathcal{J}_{ij}(x) = \sum_{k=1}^{n+1} \frac{1}{x_k} \frac{\partial x_k}{\partial x_i} \frac{\partial x_k}{\partial x_j}.$$

The Fisher information metric has several interesting properties. Its inverse is the asymptotical variance of the MLE and it corresponds to a bound on the estimation error of unbiased estimators (Cramer-Rao lower bound). Intuitively, $\det \mathcal{J}(x)$ represent the amount of information a sample point conveys with respect to the problem of estimating the parameter $x$. Perhaps the most interesting property is given by Čencov's theorem [2] which states that the Fisher information metric is uniquely determined by invariance under sufficient statistics transformations.

We now describe a well-known way of characterizing the Fisher information as a pull-back metric from the positive $n$-sphere $\mathcal{S}_n^+$ (see for example [6])

$$\mathcal{S}_n^+ = \left\{ x \in \mathbb{R}^{n+1} : \forall i, \ x_i \geq 0, \ \sum_{i=1}^{n+1} x_i^2 = 1 \right\}.$$

The transformation $R : \mathcal{P}_n \to \mathcal{S}_n^+$ defined by

$$R(x) = (\sqrt{x_1}, \ldots, \sqrt{x_{n+1}})$$

pulls-back the Euclidean metric on the surface of the sphere to the Fisher information on the multinomial simplex. In other words, the geodesic distance $d(x, y)$ for $x, y \in \mathcal{P}_n$ under the Fisher information metric may be obtained by measuring the length of the great circle on $\mathcal{S}_n^+$ between $R(x)$ and $R(y)$

$$d(x,y) = \text{acos}\left( \sum_{i=1}^{n+1} \sqrt{x_i y_i} \right).$$

Consider now the following family of diffeomorphisms $F_\lambda : \text{int}\mathcal{P}_n \to \text{int}\mathcal{P}_n$

$$F_\lambda(x) = \left( \frac{x_1 \lambda_1}{x \cdot \lambda}, \ldots, \frac{x_{n+1} \lambda_{n+1}}{x \cdot \lambda} \right), \quad \lambda \in \text{int}\mathcal{P}_n$$

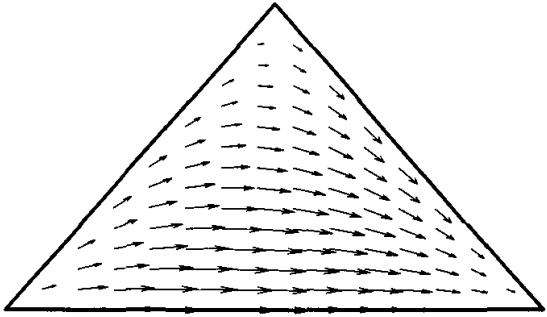

Figure 1: $F_\lambda$ acting on $\mathcal{P}_2$ for $\lambda = (\frac{2}{10}, \frac{5}{10}, \frac{3}{10})$.

where $x \cdot \lambda$ is the scalar product $\sum_{i=1}^{n+1} x_i \lambda_i$. The family $F_\lambda$ is a Lie group of transformations under composition that is isomorphic to $\text{int}\mathcal{P}_n$. The identity element is $(\frac{1}{n+1}, \ldots, \frac{1}{n+1})$ and the inverse of $F_\lambda$ is $(F_\lambda)^{-1} = F_\eta$ where $\eta_i = \frac{1/\lambda_i}{\sum_k 1/\lambda_k}$.

The above transformation group acts on $x \in \text{int}\mathcal{P}_n$ by increasing the components of $x$ with high $\lambda_i$ values while remaining in the simplex. See Figure 1 for an illustration of the above action in $\mathcal{P}_2$.

We will consider the pull-back metrics of the Fisher information $\mathcal{J}$ through the above transformation group as our parametric family of metrics

$$\mathcal{G} = \{F_\lambda^* \mathcal{J} : \lambda \in \text{int}\mathcal{P}_n\}.$$

Note that since the Fisher information itself is a pull-back metric from the sphere under the square root transformation $R$ we have that $F_\lambda^* \mathcal{J}$ is also the pull-back metric of the Euclidean metric on the surface $\mathcal{S}_n^+$ through the transformation

$$\hat{F}_\lambda(x) = \left( \sqrt{\frac{x_1 \lambda_1}{x \cdot \lambda}}, \ldots, \sqrt{\frac{x_{n+1} \lambda_{n+1}}{x \cdot \lambda}} \right), \quad \lambda \in \text{int}\mathcal{P}_n.$$

As a result of the above observation we have the following closed form for the geodesic distance under $F_\lambda^* \mathcal{J}$

$$d(x, y) = \text{acos} \left( \sum_{i=1}^{n+1} \sqrt{\frac{x_i \lambda_i}{x \cdot \lambda} \frac{y_i \lambda_i}{y \cdot \lambda}} \right). \quad (3)$$

Note the only difference between (3) and TFIDF cosine similarity measure [11] is the square root and the choice of the $\lambda$ parameters.

To apply the framework described in Section 2 to the metric $F_\lambda^* \mathcal{J}$ we need to compute the volume element given by $\sqrt{\det F_\lambda^* \mathcal{J}}$. We start by computing the Gram matrix $[G]_{ij} = F_\lambda^* \mathcal{J}(\partial_i, \partial_j)$ where $\{\partial_i\}_{i=1}^n$ is a basis for $T_x \mathcal{P}_n$ given by the rows of the matrix

$$U = \begin{pmatrix} 1 & 0 & \cdots & 0 & -1 \\ 0 & 1 & \cdots & 0 & -1 \\ \vdots & 0 & \ddots & 0 & -1 \\ 0 & 0 & \cdots & 1 & -1 \end{pmatrix} \in \mathbb{R}^{n \times n+1}. \quad (4)$$

and computing $\det G$ in Propositions 1-2 below.

**Proposition 1.** *The matrix* $[G]_{ij} = F_\lambda^* \mathcal{J}(\partial_i, \partial_j)$ *is given by*

$$G = JJ^\top = U(D - \lambda \alpha^\top)(D - \lambda \alpha^\top)^\top U^\top \quad (5)$$

*where* $D \in \mathbb{R}^{n+1 \times n+1}$ *is a diagonal matrix whose entries are* $[D]_{ii} = \sqrt{\frac{\lambda_i}{x_i}} \frac{1}{2\sqrt{\lambda \cdot x}}$ *and* $\alpha$ *is a column vector given by* $[\alpha]_i = \sqrt{\frac{\lambda_i}{x_i}} \frac{x_i}{2(\lambda \cdot x)^{3/2}}$

Note that all vectors are treated as column vectors and for $\lambda, \alpha \in \mathbb{R}^{n+1}$, $\lambda \alpha^\top \in \mathbb{R}^{n+1 \times n+1}$ is the outer product matrix $[\lambda \alpha^\top]_{ij} = \lambda_i \alpha_j$.

*Proof.* The $j$th component of the vector $\hat{F}_{\lambda *} v$ is

$$[\hat{F}_{\lambda *} v]_j = \left. \frac{d}{dt} \sqrt{\frac{(x_j + tv_j)\lambda_j}{(x + tv) \cdot \lambda}} \right|_{t=0}$$

$$= \frac{1}{2} \frac{v_j \lambda_j}{\sqrt{x_j \lambda_j} \sqrt{x \cdot \lambda}} - \frac{1}{2} \frac{v \cdot \lambda \sqrt{x_j \lambda_j}}{(x \cdot \lambda)^{3/2}}.$$

Taking the rows of $U$ to be the basis $\{\partial_i\}_{i=1}^n$ for $T_x \mathcal{P}_n$ we have, for $i = 1, \ldots, n$ and $j = 1, \ldots, n+1$,

$$[\hat{F}_{\lambda *} \partial_i]_j = \frac{\lambda_j [\partial_i]_j}{2\sqrt{x_j \lambda_j}\sqrt{x \cdot \lambda}} - \frac{\sqrt{x_j \lambda_j}}{2(x \cdot \lambda)^{3/2}} \partial_i \cdot \lambda$$

$$= \frac{\delta_{j,i} - \delta_{j,n+1}}{2\sqrt{x \cdot \lambda}} \sqrt{\frac{\lambda_j}{x_j}} - \frac{\lambda_i - \lambda_{n+1}}{2(x \cdot \lambda)^{3/2}} \sqrt{\frac{\lambda_j}{x_j}} x_j.$$

If we define $J \in \mathbb{R}^{n \times n+1}$ to be the matrix whose rows are $\{\hat{F}_* \partial_i\}_{i=1}^n$ we have

$$J = U(D - \lambda \alpha^\top).$$

Since the metric $F_\lambda^* \mathcal{J}$ is the pullback of the metric on $\mathcal{S}_n^+$ that is inherited from the Euclidean space through $\hat{F}_\lambda$ we have $[G]_{ij} = \hat{F}_{\lambda *} \partial_i \cdot \hat{F}_{\lambda *} \partial_j$ hence

$$G = JJ^\top = U(D - \lambda \alpha^\top)(D - \lambda \alpha^\top)^\top U^\top.$$

□

**Proposition 2.** *The determinant of* $F_\lambda^* \mathcal{J}$ *is*

$$\det F_\lambda^* \mathcal{J} \propto \frac{\prod_{i=1}^{n+1} (\lambda_i / x_i)}{(x \cdot \lambda)^{n+1}}. \quad (6)$$

*Proof.* We will factor $G$ into a product of square matrices and compute $\det G$ as the product of the determinants of each factor. Note that $G = JJ^\top$ does not qualify as such a factorization since $J$ is not square.

By factoring a diagonal matrix $\Lambda$, $[\Lambda]_{ii} = \sqrt{\frac{\lambda_i}{x_i}}\frac{1}{2\sqrt{x\cdot\lambda}}$ from $D - \lambda\alpha^\top$ we have

$$J = U\left(I - \frac{\lambda x^\top}{x\cdot\lambda}\right)\Lambda \qquad (7)$$

$$G = U\left(I - \frac{\lambda x^\top}{x\cdot\lambda}\right)\Lambda^2\left(I - \frac{\lambda x^\top}{x\cdot\lambda}\right)^\top U^\top. \qquad (8)$$

We proceed by studying the eigenvalues and eigenvectors of $I - \frac{\lambda x^\top}{x\cdot\lambda}$ in order to simplify (8) via an eigenvalue decomposition. First note that if $(v,\mu)$ is an eigenvector-eigenvalue pair of $\frac{\lambda x^\top}{x\cdot\lambda}$ then $(v, 1-\mu)$ is an eigenvector-eigenvalue pair of $I - \frac{\lambda x^\top}{x\cdot\lambda}$. Next, note that vectors $v$ such that $x^\top v = 0$ are eigenvectors of $\frac{\lambda x^\top}{x\cdot\lambda}$ with eigenvalue 0. Hence they are also eigenvectors of $I - \frac{\lambda x^\top}{x\cdot\lambda}$ with eigenvalue 1. There are $n$ such independent vectors $v_1, \ldots, v_n$. Since $\text{trace}(I - \frac{\lambda x^\top}{x\cdot\lambda}) = n$, the sum of the eigenvalues is also $n$ and we may conclude that the last of the $n+1$ eigenvalues is 0.

The eigenvectors of $I - \frac{\lambda x^\top}{x\cdot\lambda}$ may be written in several ways. One possibility is as the columns of the following matrix

$$V = \begin{pmatrix} -\frac{x_2}{x_1} & -\frac{x_3}{x_1} & \cdots & -\frac{x_{n+1}}{x_1} & \lambda_1 \\ 1 & 0 & \cdots & 0 & \lambda_2 \\ 0 & 1 & \cdots & 0 & \lambda_3 \\ \vdots & \vdots & \ddots & \vdots & \vdots \\ 0 & 0 & \cdots & 1 & \lambda_{n+1} \end{pmatrix} \in \mathbb{R}^{n+1 \times n+1}$$

where the first $n$ columns are the eigenvectors that correspond to unit eigenvalues and the last eigenvector corresponds to a 0 eigenvalue.

Using the above eigenvector decomposition we have $I - \frac{\lambda x^\top}{x\cdot\lambda} = V\tilde{I}V^{-1}$ and $\tilde{I}$ is a diagonal matrix containing all the eigenvalues. Since the diagonal of $\tilde{I}$ is $(1, 1, \ldots, 1, 0)$ we may write $I - \frac{\lambda x^\top}{x\cdot\lambda} = V^{|n}V^{-1|n}$ where $V^{|n} \in \mathbb{R}^{n+1 \times n}$ is $V$ with the last column removed and $V^{-1|n} \in \mathbb{R}^{n \times n+1}$ is $V^{-1}$ with the last row removed.

We have then,

$$\begin{aligned}\det G &= \det(U(V^{|n}V^{-1|n})\Lambda^2(V^{-1|n\top}V^{|n\top})U^\top) \\ &= \det((UV^{|n})(V^{-1|n}\Lambda^2 V^{-1|n\top})(V^{|n\top}U^\top)) \\ &= (\det(UV^{|n}))^2 \; \det(V^{-1|n}\Lambda^2 V^{-1|n\top}).\end{aligned}$$

Noting that

$$UV^{|n} = \begin{pmatrix} -\frac{x_2}{x_1} & -\frac{x_3}{x_1} & \cdots & -\frac{x_n}{x_1} & -\frac{x_{n+1}}{x_1} - 1 \\ 1 & 0 & \cdots & 0 & -1 \\ 0 & 1 & \cdots & 0 & -1 \\ \vdots & \vdots & \ddots & \vdots & \vdots \\ 0 & 0 & \cdots & 1 & -1 \end{pmatrix} \in \mathbb{R}^{n \times n}$$

we factor $1/x_1$ from the first row and add columns $2, \ldots, n$ to column 1 thus obtaining

$$\begin{pmatrix} -\sum_{i=1}^{n+1} x_i & -x_3 & \cdots & -x_n & -x_{n+1} - x_1 \\ 0 & 0 & \cdots & 0 & -1 \\ 0 & 1 & \cdots & 0 & -1 \\ \vdots & \vdots & \ddots & \vdots & \vdots \\ 0 & 0 & \cdots & 1 & -1 \end{pmatrix}.$$

Computing the determinant by minor expansion of the first column we obtain

$$\det(UV^{|n})^2 = \left(\frac{1}{x_1}\sum_{i=1}^{n+1} x_i\right)^2 = \frac{1}{x_1^2}. \qquad (9)$$

A somewhat lengthy but straightforward argument shows that (see [8] for a proof)

$$\det V^{-1|n}\Lambda^2 V^{-1|n\top} = \frac{x_1^2(x\cdot\lambda)^{n-1}}{4^n(x\cdot\lambda)^{2n}}\prod_{i=1}^{n+1}\frac{\lambda_i}{x_i}. \qquad (10)$$

By multiplying (10) and (9) we obtain (6). □

Figure 2 displays the inverse volume element on $\mathcal{P}_1$ with the corresponding geodesic distance from the left corner of $\mathcal{P}_1$.

Propositions 1 and 2 reveal the form of the objective function $\mathcal{O}(g, D)$. In the next section we describe a maximum likelihood estimation problem that is equivalent to maximizing $\mathcal{O}(g, D)$ and study its properties.

## 5 An Inverse-Volume Probabilistic Model on the Simplex

Using proposition 2 we have that the objective function $\mathcal{O}(g, D)$ may be regarded as a likelihood function under the model

$$p(x;\lambda) = \frac{1}{Z}(x\cdot\lambda)^{\frac{n+1}{2}}\prod_{i=1}^{n+1}x_i^{1/2} \quad x \in \mathcal{P}_n, \lambda \in \text{int}\mathcal{P}_n \qquad (11)$$

where $Z = \int_{\mathcal{P}_n}(x\cdot\lambda)^{\frac{n+1}{2}}\prod_{i=1}^{n+1}x_i^{1/2}dx$. The loglikelihood function for model (11) is given by

$$\ell(\lambda;x) = \frac{n+1}{2}\log(x\cdot\lambda) - \log\int_{\mathcal{P}_n}(x\cdot\lambda)^{\frac{n+1}{2}}\prod_{i=1}^{n+1}\sqrt{x_i}\,dx.$$

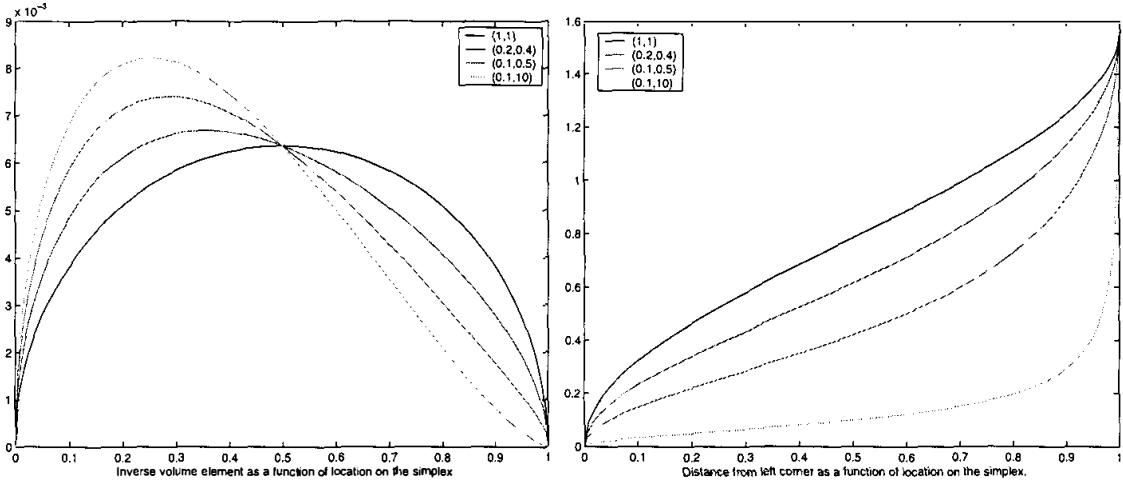

Figure 2: The inverse volume element $1/\sqrt{\det G(x)}$ as a function of $x \in \mathcal{P}_1$ (left) and the geodesic distance $d(x,0)$ from the left corner as a function $x \in \mathcal{P}_1$ (right). Different plots represent different metric parameters $\lambda \in \{(1/2, 1/2), (1/3, 2/3), (1/6, 5/6), (0.0099, 0.9901)\}$.

The Hessian matrix $H(x, \lambda)$ of the loglikelihood function may be written as

$$[H(x,\lambda)]_{ij} = -k\frac{x_i}{x \cdot \lambda}\frac{x_j}{x \cdot \lambda} - (k^2-k)L\left(\frac{x_i}{x\cdot\lambda}\frac{x_j}{x\cdot\lambda}\right) + k^2 L\left(\frac{x_i}{x\cdot\lambda}\right)L\left(\frac{x_j}{x\cdot\lambda}\right)$$

where $k = \frac{n+1}{2}$ and $L$ is the positive linear functional

$$Lf = \frac{\int_{\mathcal{P}_n}(x\cdot\lambda)^{\frac{n+1}{2}}\prod_{l=1}^{n+1}\sqrt{x_l}\ f(x,\lambda)\ dx}{\int_{\mathcal{P}_n}(x\cdot\lambda)^{\frac{n+1}{2}}\prod_{l=1}^{n+1}\sqrt{x_l}dx}.$$

Note that the matrix given by $LH(x,\lambda) = [LH_{ij}(x,\lambda)]$ is negative definite due to its covariance-like form. In other words, for every value of $\lambda$, $H(x,\lambda)$ is negative definite on average, with respect to the model $p(x;\lambda)$.

### 5.1 Computing the Normalization Term

We describe an efficient way to compute the normalization term $Z$ through the use of dynamic programming and FFT.

Assuming that $n = 2k-1$ for some $k \in \mathbb{N}$ we have

$$Z = \int_{\mathcal{P}_n} (x\cdot\lambda)^k \prod_{i=1}^{n+1} x_i^{1/2} dx$$

$$= \sum_{a_1+\cdots+a_{n+1}=k:a_i\geq 0} \frac{k!}{a_1!\cdots a_{n+1}!}\prod_{j=1}^{n+1}\lambda_j^{a_j} \int_{\mathcal{P}_n}\prod_{j=1}^{n+1}x_j^{a_j+\frac{1}{2}}$$

$$\propto \sum_{a_1+\cdots+a_{n+1}=k:a_i\geq 0}\prod_{j=1}^{n+1}\frac{\Gamma(a_j+3/2)}{\Gamma(a_j+1)}\lambda_j^{a_j}.$$

The following proposition and its proof describe a way to compute the summation in $Z$ in $O(n^2 \log n)$ time.

**Proposition 3.** *The normalization term for model (11) may be computed in $O(n^2 \log n)$ time complexity.*

*Proof.* Using the notation $c_m = \frac{\Gamma(m+3/2)}{\Gamma(m+1)}$ the summation in $Z$ may be expressed as

$$Z \propto \sum_{a_1=0}^{k} c_{a_1}\lambda_1^{a_1} \sum_{a_2=0}^{k-a_1} c_{a_2}\lambda_2^{a_2} \cdots \sum_{a_n=0}^{k-\sum_{j=1}^{n-1}a_j} c_{a_n}\lambda_n^{a_n}$$
$$c_{k-\sum_{j=1}^{n}a_j}\lambda_{n+1}^{k-\sum_{j=1}^{n}a_j}. \qquad (12)$$

A trivial dynamic program can compute equation (12) in $O(n^3)$ complexity.

However, each of the single subscript sums in (12) is in fact a linear convolution operation. By defining

$$B_{ij} = \sum_{a_i=0}^{j} c_{a_i}\lambda_i^{a_i}\cdots\sum_{a_n=0}^{j-\sum_{l=i}^{n-1}a_l} c_{a_n}\lambda_n^{a_n}c_{j-\sum_{l=i}^{n}a_l}\lambda_{n+1}^{j-\sum_{l=i}^{n}a_l}$$

we have $Z = B_{1k}$ and the recurrence relation $B_{ij} = \sum_{m=0}^{j} c_m\lambda_i^m B_{i+1,j-m}$ which is the linear convolution of $\{B_{i+1,j}\}_{j=0}^k$ with the vector $\{c_j\lambda_i^j\}_{j=0}^k$. By performing the convolution in the frequency domain filling in each row of the table $B_{ij}$ for $i = 0,\ldots,n+1, j = 0,\ldots,k$ takes $O(n\log n)$ complexity leading to a total of $O(n^2 \log n)$ complexity. □

The computation method described in the proof may be used to compute the partial derivative of $Z$, resulting in $O(n^3 \log n)$ computation for the gradient. By careful dynamic programming, the gradient vector may be computed in $O(n^2 \log n)$ time complexity as well.

## 6 Application to Text Classification

In this section we describe applying the metric learning framework to document classification and report some results on the WebKB dataset [3].

We map documents to the simplex by multinomial MLE or MAP estimation. This common representation is known as the TF (term frequency) representation and enables us to apply the geometrical structure on the multinomial simplex to documents.

It is a well known fact that less common terms across the text corpus tend to provide more discriminative information than the most common terms. In the extreme case, stopwords like the, or and of are often severely downweighted or removed from the representation. Geometrically, this means that we would like the geodesics to pass through corners of the simplex that correspond to sparsely occurring words, in contrast to densely populated simplex corners such as the ones that correspond to the stopwords above. To account for this in our framework we learn the metric $F_\lambda^* \mathcal{J} = (F_\theta^{-1})^* \mathcal{J}$ where $\theta$ is the MLE under model (11). In other words, we are pulling back the Fisher information metric through the inverse to the transformation that maximizes the normalized inverse volume of $D$.

The standard TFIDF representation of a document consists of multiplying the TF parameter by an IDF component

$$IDF_k = \log \frac{N}{\#\text{documents that word } k \text{ appears in}}.$$

Given the TFIDF representation of two documents, their cosine similarity is simply the scalar product between the two normalized TFIDF representations [11]. Despite its simplicity the TFIDF representation leads to some of the best results in text classification and information retrieval and is a natural candidate for a baseline comparison due to its similarity to the geodesic expression.

A comparison of the top and bottom terms between the metric learning and IDF scores is shown in Figure 3. Note that both methods rank similar words at the bottom. These are the most common words that often carry little information for classification purposes. The top words however are completely different for the two schemes. Note the tendency of TFIDF to give high scores to rare proper nouns while the metric learning method gives high scores for rare common nouns. This difference may be explained by the fact that IDF considers appearance of words in documents as a binary event while the metric learning looks at the number of appearances of a term in each document. Rare proper nouns such as the high scoring TFIDF terms in Figure 3 appear several times in a single web page. As a result, these words will score higher with the TFIDF scheme but lower with the metric learning scheme.

| TFIDF | Estimated $\lambda$ |
|---|---|
| tiff romano potra | disobedience seat alr |
| anitescu papeli theo | seizure refuse delegated |
| echo chimera trestle | soverigns territory |
| schlatter xiyong | mobocracy stabbed |
| $\vdots$ | $\vdots$ |
| at department with | will course system |
| this by office course | you page research with |
| are an from system | that by are at this |
| programming be last | home from office or as |

Figure 3: Comparison of top and bottom valued parameters for TFIDF and model (11). The dataset is the faculty vs. student webpage classification task from WebKB dataset. Note that the least scored terms are similar for the two methods while the top scored terms are completely disjoint.

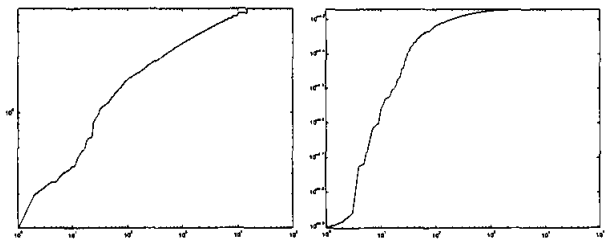

Figure 4: Log-log plots for sorted values of TFIDF (top) and estimated $\lambda$ values (bottom). The task is the same as in Figure 3.

In Figure 4 the rank-value plot for the estimated $\lambda$ values and IDF is shown on a log-log scale. The $x$ axis represents different words that are sorted by increasing parameter value and the $y$ axis represents the $\lambda$ or IDF value. Note that the IDF scores show a much stronger linear trend in the log-log scale than the $\lambda$ values.

To measure performance in classification we compared the testing error of a nearest neighbor classifier under several different metrics. We compared TFIDF cosine similarity, $L_2$ distance for TF representation and the geodesic distance under the metric obtained by the inverse transformation to the MLE. Figure 5 displays test-set error rates as a function of the training set size. The error rates were averaged over 20 experiments with random sampling of the training set. The $\lambda$ parameter was obtained by gradient descent using the dynamic programming method described in Section 5.1. According to Figure 5 the method described in this paper outperforms the two alternatives.

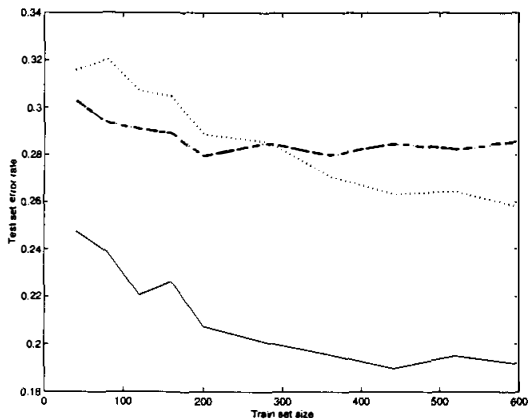

Figure 5: Test set error rate for nearest neighbor classifier on the WebKB faculty vs. student task. Distances were computed by geodesic for the learned Riemannian metric (solid), TFIDF with cosine similarity (dashed) and TF with $L_2$ norm (dotted).

## 7 Summary

We have proposed a new framework for the metric learning problem that enables robust learning of a local metric for high dimensional sparse data. This is achieved by restricting the set of metric candidates to a parametric family and selecting a metric based on maximizing the inverse volume element.

In the case of learning a metric on the multinomial simplex, the metric candidates are taken to be pull-back metrics of the Fisher information under a continuous group of transformation. When composed with a square root, the transformations are flattening transformation for the obtained metrics. The resulting optimization problem may be interpreted as maximum likelihood estimation.

Guided by the well known principle that common words should have little effect on the metric structure we learn the metric that is associated with the inverse to the transformation that maximizes the inverse volume of the training set. The resulting pull-back metric de-emphasizes common words, in a way similar to TFIDF. Despite the similarity between the resulting geodesics and TFIDF similarity measure, there are significant qualitative and quantitative differences between the two methods. Using a nearest neighbour classifier in a text classification experiment, the obtained metric is shown to significantly outperform other metrics such as TFIDF cosine similarity and a TF based $L_2$ distance.

The framework proposed in this paper is quite general and allows implementations in other domains. The key component is the specification of the set of metric candidates possibly by parametric transformations.

### Acknowledgements
I thank John Lafferty and Leonid Kontorovich for interesting discussions and helpful comments.